\documentclass[10pt,twocolumn,letterpaper]{article}

\usepackage[pagenumbers]{cvpr} 

\usepackage{graphicx}
\usepackage{amsmath}
\usepackage{amssymb}
\usepackage{booktabs}
\usepackage{paralist}

\usepackage[pagebackref,breaklinks,colorlinks]{hyperref}

\usepackage[capitalize]{cleveref}

\newcommand{\squeezeup}{\vspace{-2.5mm}}

\crefname{section}{Sec.}{Secs.}
\Crefname{section}{Section}{Sections}
\Crefname{table}{Table}{Tables}
\crefname{table}{Tab.}{Tabs.}

\def\cvprPaperID{*****} 
\def\confName{CVPR}
\def\confYear{2022}

\begin{document}

\title{Towards Optimizing OCR for Accessibility}

\author{Peya Mowar \hspace{10pt} Tanuja Ganu \hspace{10pt} Saikat Guha\\
Microsoft Research, India \\
{\tt\small \{t-peyamowar, tanuja.ganu, saikat\}@microsoft.com}}
\maketitle

\begin{abstract}
Visual cues such as structure, emphasis, and icons play an important role in efficient information foraging by sighted individuals and make for a pleasurable reading experience.  Blind, low-vision and other print-disabled individuals miss out on these cues since current OCR and text-to-speech software ignore them, resulting in a tedious reading experience. We identify four semantic goals for an enjoyable listening experience, and identify syntactic visual cues that help make progress towards these goals. Empirically, we find that preserving even one or two visual cues in aural form significantly enhances the experience for listening to print content.
\end{abstract}

\section{Introduction}
\label{sec:intro}



Worldwide, there are over 250 million blind and low-vision (BLV), or otherwise print-disabled individuals~\cite{some-report}. Accessible long-form content, such as textbooks, magazines, newspapers, and literature, is a key enabler for their intellectual enrichment and high quality of life. Braille and audio books are scarce or expensive, while human-mediated solutions like having friends and family read out content is an imposition reserved only for the most critically needed content.

\begin{figure}[t]
\centering
\includegraphics[width=1\linewidth]{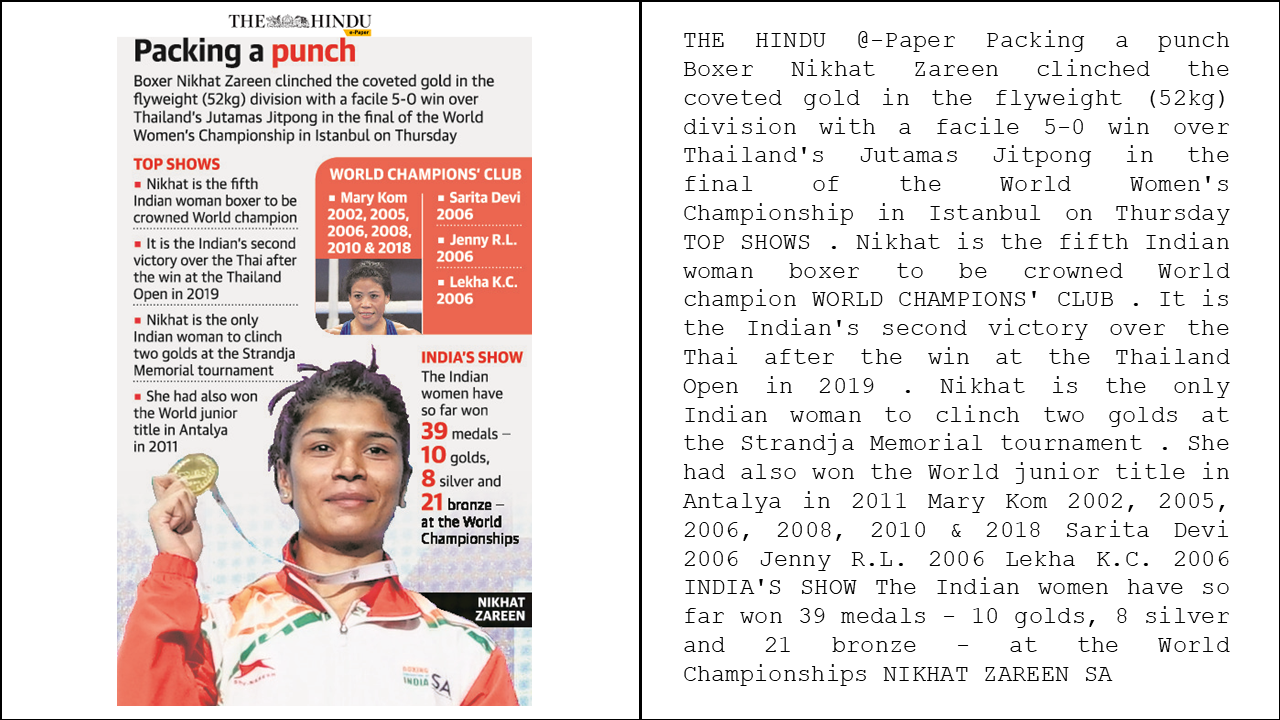}
\squeezeup
\caption{A news article and it's OCR output.}
\squeezeup
\label{fig:ocr}
\end{figure}

Computer vision based apps~\cite{lens} that combine optical character recognition (OCR) and text-to-speech (TTS) tend to be used for micro-content, such as labels on medicines and packaged goods, not for long-form content. Figure~\ref{fig:ocr} illustrates why. On left is an article from a daily newspaper, and on the right is the OCR output of the same from a popular OCR app. The OCR output lacks all the visual cues (e.g., structure, emphasis, icons) that sighted users depend on to get a quick overview of the content before selectively drilling down. The lack of visual cues combined with the linear sequential nature of audio ruins the enjoyability of automatically converted long-form content.

OCR was never optimized for the accessibility use case. Driven primarily by the information retrieval use-case, whether searching in books~\cite{gbooks} or recognizing handwritten postal codes~\cite{mnist}, OCR entirely ignores visual cues as evidenced by the output of popular OCR libraries and services~\cite{pytesseract, googleocr, acs}. At the same time, recognizing all possible visual cues across all languages is a significant undertaking with questionable utility for the accessibility use case given the constraints placed by the linearity and limited aural cues (e.g. voice, tempo) of the eventual audio output. 

This work argues for more research in preserving visual cues in aural form and takes the first steps in that direction. Specifically, we \begin{inparaenum}[i)]
\item define the OCR for accessibility problem and propose metrics for \textit{glanceability}, \textit{readability}, \textit{listenability}, and \textit{satisfiability} of long-form content,
\item identify the simplest visual cues that, when preserved as aural cues, significantly enhance the accessibility of long-form content, and
\item apply our approach to daily newspaper content and provide initial results.
\end{inparaenum}
Overall, we believe a concerted effort into optimizing OCR and TTS for accessibility of long-form content is possible and holds significant promise in enriching the lives of print-disabled individuals.   

\section{Background}
\label{sec:background}

\subsection{Accessibility Research} 

Accessibility research has majorly concentrated on increasing digital access for BLV users\cite{10.1145/3411764.3445412}. This includes improving the narrative experience for digital mediums such as IDEs\cite{10.1145/3411764.3445544, 10.1145/3173574.3174192}, collaborative documents\cite{10.1145/3491102.3517635} and scientific reports\cite{https://doi.org/10.48550/arxiv.2105.00076}. The focal point of the research community has been generating non-visual representations for visual content. This is done either in the form of rich and descriptive alt-text\cite{10.1145/3173574.3173633} or using auditory cues such as earcons and their derivatives (spearcons, lyricons, etc.)\cite{10.1145/169059.169179}.

These non-visual representations are generated to substitute ``visual information scents" that help users easily navigate to information of their interest. This process is referred to as information foraging and has been studied extensively for web access to BLV users\cite{10.1145/3334480.3383025}.

\subsection{Text Recognition and OCRs} 
Optimal character recognition (popularly known as OCR) systems and more broadly text recognition systems extract text from images. They are mostly utilized as a form of data entry for digitizing print content such as bank statements, legal records, healthcare reports, and invoices. They are also employed in applications such as machine translation\cite{glens} and cognitive computing\cite{azure}.

Table \ref{tab:ocr} summarizes the output functionalities of some state-of-the-art OCR APIs. The research in this domain contributes in improving these OCRs in one of three ways: \begin{inparaenum}[i)] 
\item scene-text, handwriting, cursive text recognition (in the wild) \cite{8099766, 8953886, 9157011, Qiao_Tang_Cheng_Xu_Niu_Pu_Wu_2020, 9157479, Wan_He_Chen_Bai_Yao_2020}
\item developing models for Indic languages\cite{Dwivedi_2020_CVPR_Workshops} and 
\item extracting text in the correct reading order\cite{Tang_Sun_Jin_Zhang_2021, wang2021layoutreader, lee-etal-2021-rope}.
\end{inparaenum} 

\begin{table*}[h]
  \centering
  \resizebox{0.8\linewidth}{!}{
  \begin{tabular}{@{}l|c@{}}
    \toprule
    \bf{OCR APIs} & \bf{Output Functionalities} \\
    \midrule
    PyTesseract & Recognised characters, box boundaries, confidences, orientation, script \\
    Azure Cognitive Services Read API & Recognised characters, words, sentences, box boundaries, confidences, orientation, script, whether handwriting \\
    Google Cloud OCR & Recognised words, phrases, polygonal boundaries, confidences, script\\
    \bottomrule
  \end{tabular}
  }
  \squeezeup
  \caption{Output functionalities of state-of-the-art OCR APIs}
  \squeezeup
  \label{tab:ocr}
\end{table*}


\subsection{Research Gaps} 
Prior accessibility research has not focused on print-impaired users (that include BLV, illiterate, and users with learning and physical disabilities), and more specifically on print accessibility. More work has been done for digital access, where text metadata is more commonly available in the form of HTML tags and alternate texts. Non-profits such as the DAISY Forum \cite{daisy} in India have addressed this need, however, they rely on manual intervention due to the lack of important metadata in the OCR text outputs. This leads to a huge restraint in the speed and costs associated with digitization for the print-impaired.

The required metadata is nothing else but the proxy for the visual information scents present in printed content. These visual cues or “syntaxes” include but are not limited to bold, italics, underline, highlighting, font style, size, color, ordered and unordered lists, shapes such as boxes and lines, spacing, indentation, alignment, shadow, background color. Since current OCR systems lack in providing this information, a print-disabled reader is unable to leverage these cues resulting in a higher time cost to gain value from the same content.

Thus, there is a need for text recognition systems that can preserve these visual cues for narration. However, not all syntaxes may be equally relevant to optimize the users' information gain, and thus, it is vital to understand the degree of abstraction necessary to alleviate the accessibility challenges presented by their underlying semantics\cite{loukina-etal-2018-towards}.

\section{Methodology}
\label{sec:methodology}

Our research goal is to identify the visual cues, which when presented to the users, result in an optimal rate of information gain. To do this, we consider the underlying semantics of these cues, the lack of which pose certain accessibility challenges. We classify these challenges in long-form text content into four broad categories:
\begin{inparaenum}

\begin{figure}[h]
\centering
\squeezeup
\includegraphics[width=0.8\linewidth]{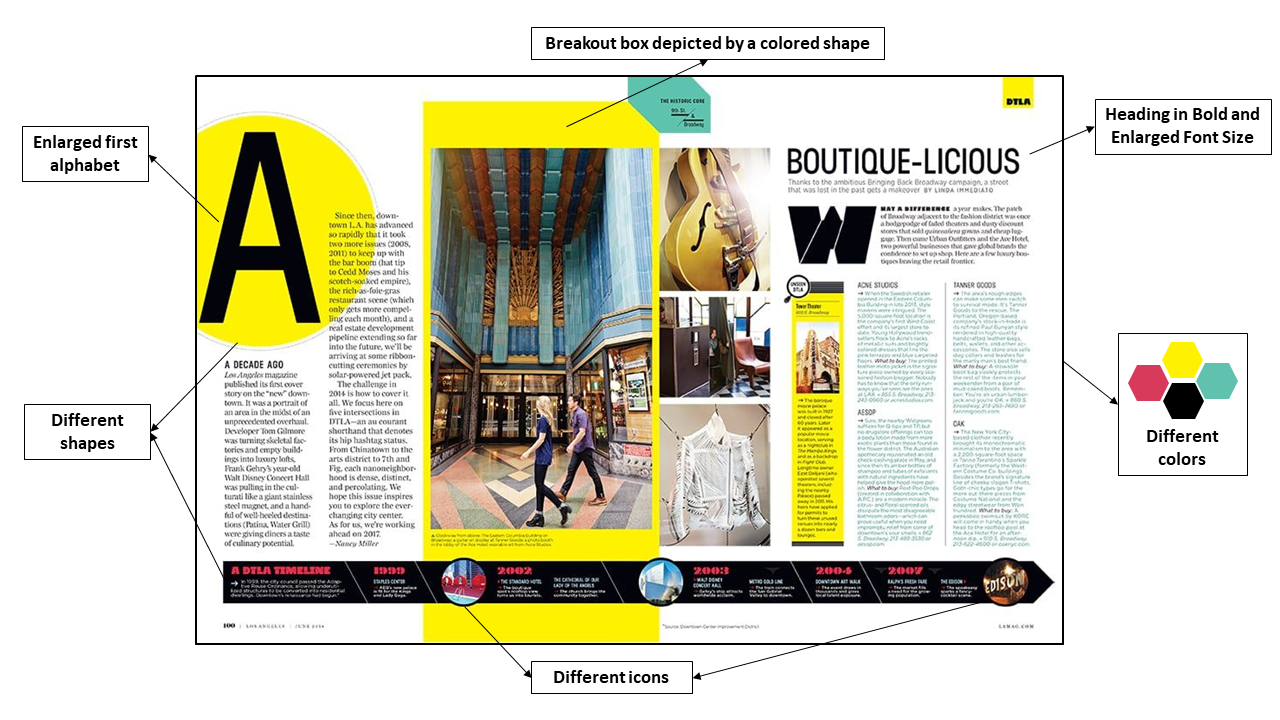}
\squeezeup
\caption{Visual cues in a page from a magazine.}
\squeezeup
\label{fig:magazine}
\end{figure}

\item \textbf{\textit{Glanceability}}: 
In order to emphasise key content for the users, long-form texts such as newspapers, books and magazines contain content with varying grades of emphasis. For example, breaking news is often centered on the first page, occupying a significant amount of area to grab the reader’s attention; and similarly, headings and subheadings in books or magazines often have bolden text and a larger font size than the remaining text as shown in Figure~\ref{fig:magazine}. This relies on the users' ability to glance at the page, view the emphasized content and get the gist of the whole content, and is a challenge for print-impaired users who do not get access to the emphasized content from an OCR. 

\item \textbf{\textit{Readability}}: 
This is the ease with which the users can comprehend the content. Sighted users can understand the organization and structure of the content due to the text formatting such as a breakout box depicted using a colored shape in the background or list elements marked by bullet points or enlarged first alphabets as shown in Figure~\ref{fig:magazine}. On the other hand, print-impaired users listen to the same content with no context being provided about the structure of the content, which is innately visual in nature and thus lose out on the comprehensibility of the content.  

\item \textbf{\textit{Listenability}}: 
The listening experience very often differs from the reading experience of content due to the lack of spatial information which can be randomly accessed. Revealing too many visual cues in the narration may become overwhelming for the listener. Thus, along with \textit{readability}, it is also fitting to take \textit{listenability} of the content into consideration. This is done by considering the ability of the users to listen, understand and enjoy the narrative version of a long-form text content. 

\item \textbf{\textit{Satisfiability}}:
Often in print media, a plethora of colors, icons, shapes, and font styles are used, as shown in Figure~\ref{fig:magazine}, in order to attract and retain the readers, improve their overall reading experience and most importantly, make it fun and create appropriate psychological impact. Thus, when we digitize the same print content for print-impaired users, it becomes imperative to take these fun visual elements into consideration and make the narration a rewarding and enjoyable experience for them. This is what we refer to as \textit{satisfiability} - when the content meets the user expectations. 
\end{inparaenum}

\section{Experiments and Learnings}
\label{sec:experiments}


We conducted two sets of experiments - an interview study and a survey to find the least degree of abstractive formatting required that can address these challenges defined in Section \ref{sec:methodology}. We perform our initial set of experiments on 8 articles extracted from the front page of \textbf{The Hindu}, a popular English daily newspaper in India. The articles contain 250-500 words and a diverse set of visual cues such as breakout boxes, list elements and figures.

\subsection{Interview Study}

The objective of this experiment was to observe patterns in the reading behaviour of sighted users and derive insights on \textit{glanceability}, \textit{readability} and \textit{satisfiability}.

\subsubsection{Experiment Setup}
We conducted task-based, semi-structured user interviews. We recruited 8 participants for reading the content. All participants were above 18 years of age, sighted and proficient in the English language. The interviews were conducted over video conferencing and were 30 minutes long in duration.

For each article, we extracted the text using the Azure Read API ~\cite{acs} and generated four stages of visual formatting, as explained in Table~\ref{tab:formats}. We created four decks - each containing the 8 articles, 2 articles per a format style. The participants were randomly allotted one deck to perform their tasks on.

\begin{table*}[h]
  \centering
  \resizebox{0.8\linewidth}{!}{
  \begin{tabular}{@{}c|p{7in}@{}}
    \toprule
    \bf{Visual Formats} & \bf{Description}\\
    \midrule
    F0 & Plain text extracted from the article using Azure Cognitive Services OCR in ‘natural reading order’ \\
    F1 & To F0, added emphasis by boldening the headline, subheading, byline, breakout box as well as metadata \\
    F2 & To F1, changing font colors for all the text present in bounding boxes to provide a sense of structuring\\
    F3 & Original article image with all formatting intact\\
    \toprule
    \bf{Auditory format} & \bf{Description} \\
    \midrule
    A0 & F0 narrated as is by the voice ‘Neerja – English Indian’ \\
    A1 & F1 narrated where the bolden text narrated by the voice ‘Prabhat – English Indian’ and the plain article text narrated by ‘Neerja – English Indian’\\
    \bottomrule
  \end{tabular}
  }
  \squeezeup
  \caption{Considered visual and auditory formats}
  \squeezeup
  \label{tab:formats}
\end{table*}

\subsubsection{Evaluation} 
\begin{inparaenum}
\item \textbf{\textit{Glanceability}}: We evaluated \textit{glanceability} by noting the time taken by the users in identifying the key content from a given text content. In the interview, we requested the participants to find and locate all the subheadings present in the news article. We then noted the the accuracy and the time that the participants took in finding these subheadings.

\item \textbf{\textit{Readability}}: We measured the time taken by the users in identifying different structural elements to evaluate \textit{readability} of the content. In our interviews, the participants were asked to identify if the news article contains a breakout box (which typically contains a summarization, timeline, quote or statistics) in the article. We noted the time and accuracy with which the participants responded.

\item \textbf{\textit{Satisfiability}}: The metric we used to evaluate \textit{satisfiability} was the users’ self-assessment of their reading experience based on the User Engagement Scale ~\cite{OBRIEN201828}. We asked the users to rate their responses to the following two statements on a Likert Scale ranging from Strongly Disagree to Strongly Agree (1-5): \begin{inparaenum}[i)]
\item The reading experience was rewarding.
\item The reading experience was enjoyable.
\end{inparaenum}
\end{inparaenum}
\vspace{-9pt}
\subsubsection {Quantitative Results}
\begin{inparaenum}
\item \textbf{\textit{Glanceability}}: 
\textit{Accuracy} - Without any visual cues (F0), the participants were able to correctly identify the subheadings only 18.75\% of the time. Simply by boldening the important text (F1), the accuracy raised to 68.75\%. \textit{Time Cost} - On an average, the participants spent 43 seconds to identify the headline in unformatted text (F0). In fact, 43.75\% of the time, the users spent less than 30 seconds and gave up. Moreover, of the users who gave accurate responses for the task, the median time spent to find the subheading in unformatted text was 80 seconds, significantly higher than any of the other formats, as is evident from Figure~\ref{fig:bar-chart1}.

\item \textbf{\textit{Readability}}: 
\textit{Accuracy} - The accuracy of responses positively correlated with the presence of visual cues, rising from 25\% when no cues were given (F0) to 81.25\% when text was emphasised and structured (F2). \textit{Time Cost} - The median time spent by the participants on unformatted text was 91 seconds, and this time dropped significantly (to 20 seconds) with a single grade increase in formatting (F1). This is illustrated in Figure~\ref{fig:bar-chart2}.

\begin{figure}[h]
  \centering
  \begin{subfigure}{0.48\linewidth}
    \includegraphics[width=1\linewidth]{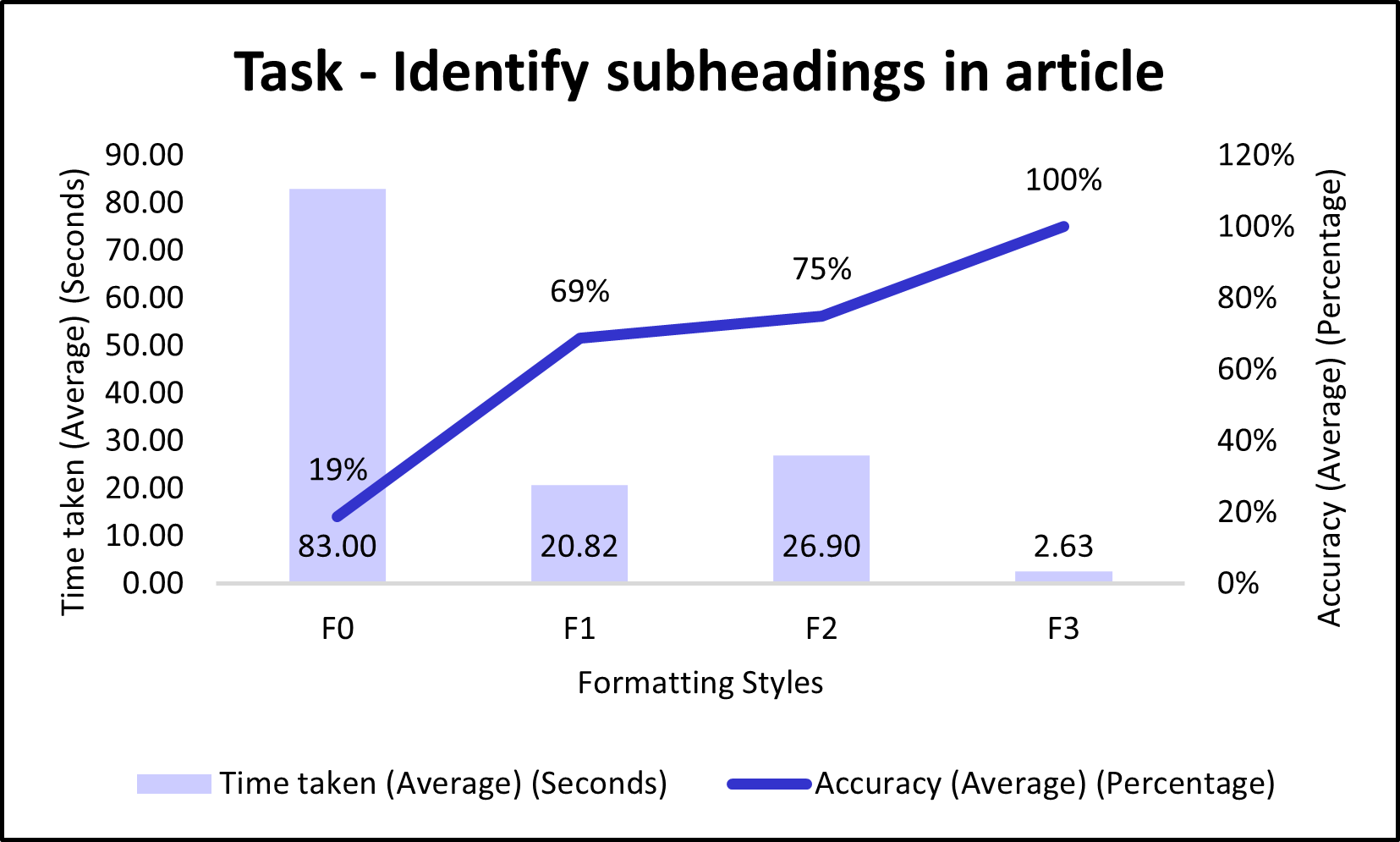}
    \caption{Glanceability.}
    \label{fig:bar-chart1}
  \end{subfigure}
  \hfill
  \begin{subfigure}{0.48\linewidth}
    \includegraphics[width=1\linewidth]{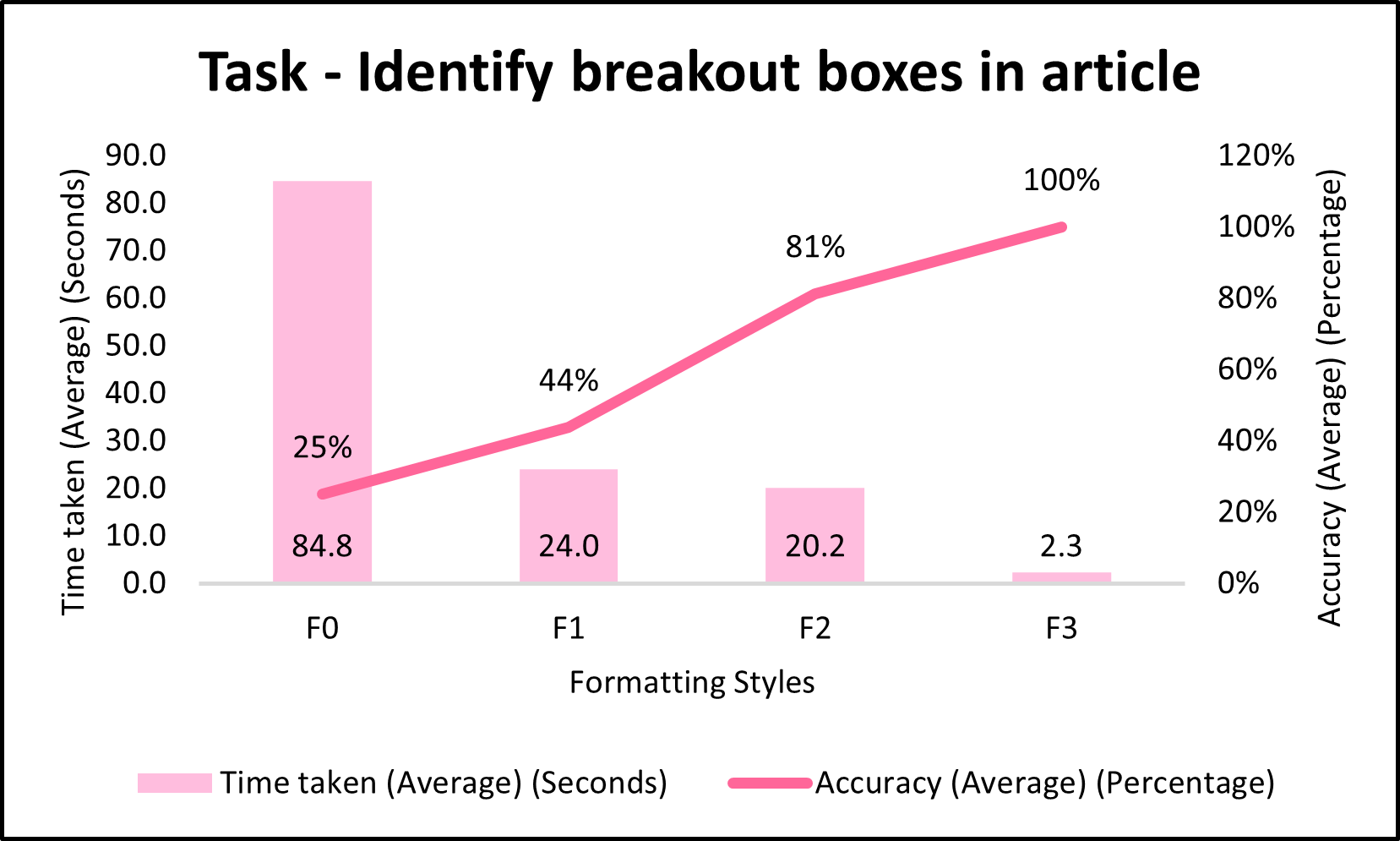}
    \caption{Readability.}
    \label{fig:bar-chart2}
    \end{subfigure}
  \centering
  \begin{subfigure}{0.6\linewidth}
    \includegraphics[width=1\linewidth]{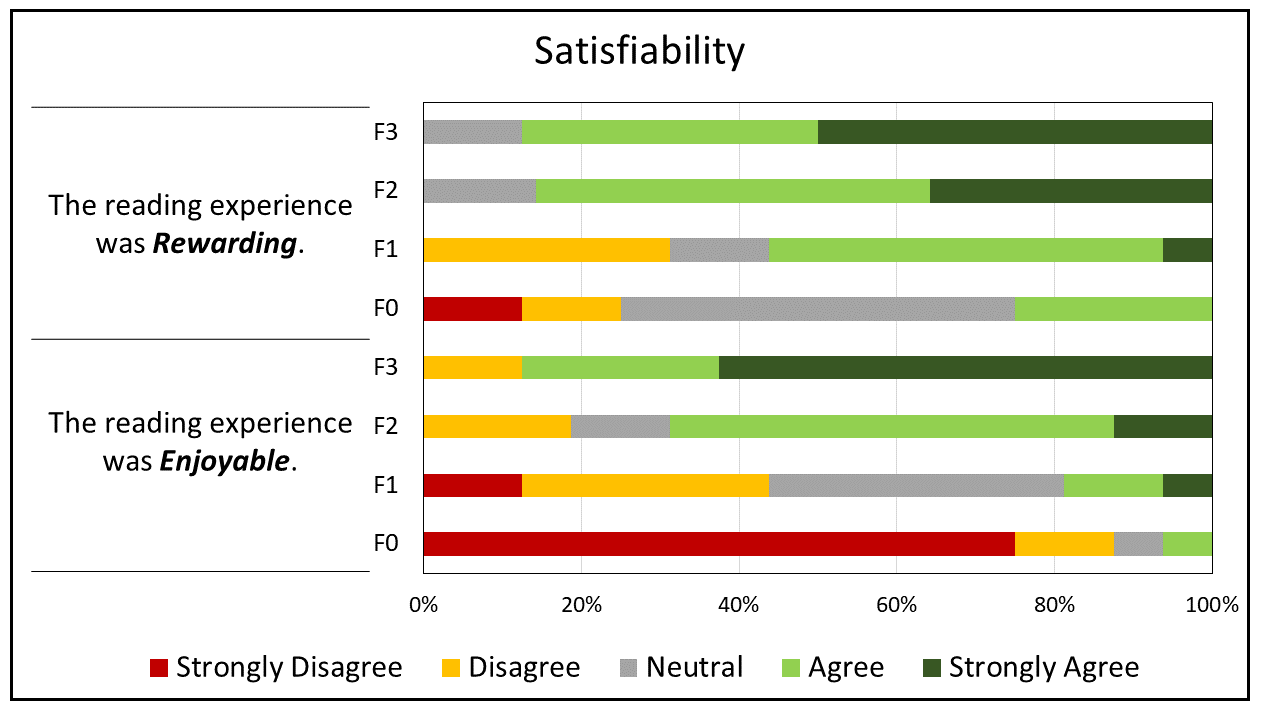}
    \caption{Satisfiability.}
    \label{fig:stacked-bar-chart1}
  \end{subfigure}
  \vfill
  \begin{subfigure}{0.6\linewidth}
    \includegraphics[width=1\linewidth]{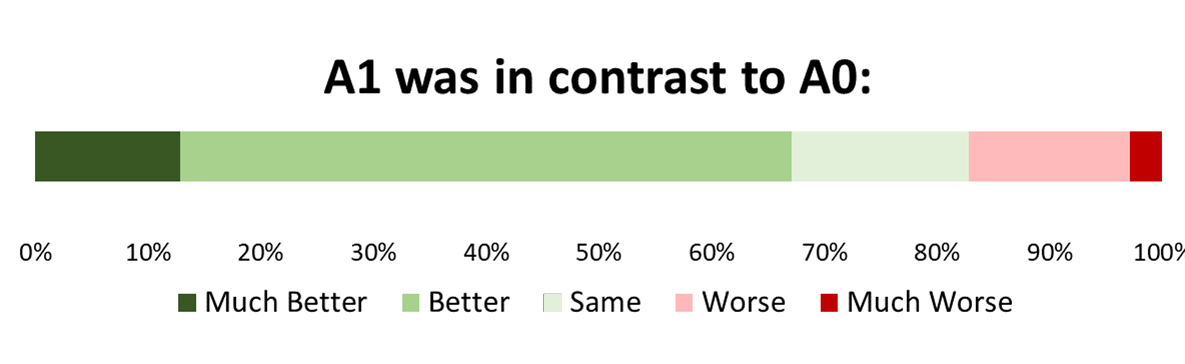}
    \caption{Listenability.}
    \label{fig:stacked-bar-chart2}
  \end{subfigure}
  \squeezeup
   \caption{Visualisations of the quantitative results.}
   \squeezeup
\end{figure}

\item \textbf{\textit{Satisfiability}}: 
On an average, we found that participants found the content to be both more \textit{rewarding} and \textit{enjoyable} with more visual cues. This is illustrated in Figure~\ref{fig:stacked-bar-chart1}.
\end{inparaenum}


\subsection{Survey}
The objective of this experiment was to garner insights on the relative \textit{listenability} of two audios and validate if inserting the simplest aural cues in the narration improves the overall listening experience of the users.
\vspace{-10pt}
\subsubsection{Experiment Setup}

We distributed the survey to our friends and family and ended up receiving 12 responses. All participants were above 18 years of age, sighted and proficient in the English language. The survey took an average time of 40 minutes for completion.

We leveraged the Azure Speech Studio~\cite{speech} for converting two of the previously mentioned formats to audio narrations. The audio formats are explained further in detail in Table~\ref{tab:formats}.

\subsubsection{Evaluation and Quantitative Results}

\begin{inparaenum}
\item \textbf{\textit{Metrics}}: We evaluated \textit{listenability} by asking the users to compare between two audio clips of different formats and answer whether one was `much better', `better', `same', `worse' or `much worse' than the other. Moreover, we also took the listener's qualitative inputs on their listening experience into account.

\item \textbf{\textit{Results}}: We discovered that simply by emphasizing (A1), the narration became 31\% better (than A0) for the listeners. Figure~\ref{fig:stacked-bar-chart2} illustrates the results of the survey.
\end{inparaenum}


\subsection{Qualitative Results}
\begin{inparaenum}
\item \textit{Heavy reliance on bolden text}: We observed that some participants completely relied on bolden text to find the subheadings. P2, when presented with unformatted text (F0), simply gave up the search:
\textit{"This article doesn't have any subheadings. There's nothing in bold"}. Moreover, some participants declared that everything bolden in F1 (headline, meta-data, byline, etc.) was the subheading, without actually going through the content.

\item \textit{Cases and Punctuation as substitutes}: During the interviews, participants heavily depended on cases and punctuation (visual cues picked up by the OCR) for finding elements. Text in quotes were declared as subheadings, and a continuous stream of uppercase letters as breakout boxes. This had a negative effect. For instance, an article had a quote inside a breakout box. No participant was able to correctly identify that breakout box.

\item \textit{Emphasis provided a break in monotony}: The survey participants pointed out that usage of two different voices helped break the monotonous flow of the audio, provided a better outline of the article and improved the overall listening experience - \textit{"The variation ( the use of a different voice to speak out, what mostly appeared to be heading), helping alleviating the monotonicity of the spoken content."}.

\item \textit{Lack of visual appeal affected reading experience}: Some participants were severely underjoyed to see unformatted text (F0) and rated the reading experience without actually going through it in entirety. This points to how appeal plays a major role in attracting a reader and making it a rewarding experience. When in the deck, the original article (F3) showed up, the reaction was of pure bliss and comfort. 

\end{inparaenum}


\section{Conclusion}

In this paper, we articulated the need for preserving visual cues in aural form in order to enhance the accessibility of printed content for blind, low-vision, and other print-disabled individuals. We found that existing OCR services and current research in OCR largely ignore visual cues that are necessary for information foraging. We proposed metrics for \textit{glanceability}, \textit{readability}, \textit{listenability} and overall \textit{satisfiability}, and characterised how adding even one or two visual cues in aural form significantly improve the narration experience for print-disabled individuals.

{\small
\bibliographystyle{ieee_fullname}
\bibliography{egbib}
}

\end{document}